\documentclass[11pt,a4paper]{article}
\usepackage[hyperref]{emnlp2018}
\usepackage{times}
\usepackage{latexsym}
\usepackage{graphicx}
\usepackage{array}
\usepackage{url}
\usepackage{multirow}
\usepackage{multicol}
\usepackage{changes}
\definechangesauthor[color=red]{lyf}
\definechangesauthor[color=blue]{luqin}

\makeatletter
\newcommand{\thickhline}{%
    \noalign {\ifnum 0=`}\fi \hrule height 2pt
    \futurelet \reserved@a \@xhline
}
\newcolumntype{"}{@{\hskip\tabcolsep\vrule width 1pt\hskip\tabcolsep}}
\makeatother

\aclfinalcopy 
\def\aclpaperid{***} 

\title{Dual Memory Network Model for Biased Product Review Classification}

\author{Yunfei Long\textsuperscript{1}\textsuperscript{*}, Mingyu Ma\textsuperscript{1}\textsuperscript{*}, Qin Lu\textsuperscript{1}, Rong Xiang\textsuperscript{1} \and Chu-Ren Huang\textsuperscript{2}\\
  {\textsuperscript{1}Department of Computing, The Hong Kong Polytechnic University}\\
  {{csylong,csluqin,csrxiang}@comp.polyu.edu.hk, derek.ma@connect.polyu.hk}\\
  {\textsuperscript{2}Department of Chinese and Bilingual Studies, The Hong Kong Polytechnic University}\\
  {\textsuperscript{*}These two authors contributed equally}\\
  {churen.huang@polyu.edu.hk}}

\date{}

\begin{document}
\maketitle
\begin{abstract}
In sentiment analysis (SA) of product reviews, both user and product information are proven to be useful. Current tasks handle user profile and product information in a unified model which may not be able to learn salient features of users and products effectively.  In this work, we propose a dual user and product memory network (DUPMN) model to learn user profiles and product reviews using separate memory networks. Then, the two representations are used jointly for sentiment prediction. The use of separate models aims to capture user profiles and product information more effectively. Compared to state-of-the-art unified prediction models, the evaluations on three benchmark datasets, IMDB, Yelp13, and Yelp14, show that our dual learning model gives performance gain of 0.6\%, 1.2\%, and 0.9\%, respectively. The improvements are also deemed very significant measured by \textit{p-values}. 
\end{abstract}

\section{Introduction}
Written text is often meant to express sentiments of individuals. Recognizing the underlying sentiment expressed in the text is essential to understand the full meaning of the text. The SA community is increasingly interested in using natural language processing (NLP) techniques as well as sentiment theories to identify sentiment expressions in the text. 

Recently, deep learning based methods have taken over feature engineering approaches to gain further performance improvement in SA. Typical neural network models include Convolutional Neural Network (CNN) \cite{kim2014convolutional}, Recursive auto-encoders \cite{socher2013recursive}, Long-Short Term Memory (LSTM) \cite{tang2015document}, and many more. 

Attention-based models are introduced to highlight important words and sentences in a piece of text. Different attention models are built using information embedded in the text including users, products and text in local context \cite{tang2015learning,yang2016hierarchical,chen2016neural,gui2016intersubjectivity}. In order to incorporate other aspects of knowledge, Qian et al. \shortcite{qian2016linguistically} developed a model to employ additional linguistic resources to benefit sentiment classification. Long et al.\shortcite{long2017cognition} and Mishra et al.\shortcite{mishra2016leveraging} proposed cognition-based attention models learned from cognition grounded eye-tracking data. 

Most text-based SA is modeled as sentiment classification tasks. In this work, SA is for product reviews. We use the term \textbf{users} to refer to writers of text, and \textbf{products} to refer to the targets of reviews in the text. A \textbf{user profile} is defined by the collection of reviews a user writes. \textbf{Product information} defined for a product is the collection of reviews for this product. Note that user profiles and product information are not independent of each other. That is one reason why previous works use unified models. By commonsense we know that review text written by a person may be subjective or biased towards his/her own preferences. Lenient users tend to give higher ratings than finicky ones even if they review the same products. Popular products do receive higher ratings than those unpopular ones because the aggregation of user reviews still shows the difference in opinion for different products. While users and products both play crucial roles in sentiment analysis, they are fundamentally different. 

Reviews written by a user can be affected by user preference which is more subjective whereas reviews for a product are useful only if they are from a collection of different reviewers, because we know individual reviews can be biased. The popularity of a product tends to reflect the general impression of a collection of users as an aggregated result. Therefore, sentiment prediction of a product should give dual consideration to individual users as well as all reviews as a collection.

In this paper, we address the aforementioned issue by proposing to learn user profiles and product review information separately before making a joint prediction on sentiment classification. In the proposed Dual User and Product Memory Network (DUPMN) model, we first build a hierarchical LSTM \cite{hochreiter1997long} model to generate document representations. Then a user memory network (UMN) and a product memory network (PMN) are separately built based on document representation of user comments and product reviews. Finally, sentiment prediction is learned from a dual model.

To validate the effectiveness of our proposed model, evaluations are conducted on three benchmarking review datasets from IMDB and Yelp data challenge (including Yelp13 and Yelp14) \cite{tang2015document}. Experimental results show that our algorithm can outperform baseline methods by large margins. Compared to the state-of-the-art method, DUPMN made 0.6\%, 1.2\%, and  0.9\% increase in accuracy with \textit{p-values} 0.007, 0.004, and 0.001 in the three benchmark datasets respectively. Results show that leveraging user profile and product information separately can be more effective for sentiment predictions.

The rest of this paper is organized as follows. Section \ref{sec:Related works} gives related work, especially memory network models. Section \ref{sec:proposaledmodel} introduces our proposed DUPMN model. Section \ref{sec:Experiment} gives the evaluation compared to state-of-the-art methods on three datasets. Section \ref{sec:Conclusion and future works} concludes this paper and gives some future directions in sentiment analysis models to consider individual bias.

\section{Related Work}\label{sec:Related works}
Related work includes neural network models and the use of user/product information in sentiment analysis.

\subsection{Neural Network Models}\label{subsec:neuralmodel} 
In recent years, deep learning has greatly improved the performance of sentiment analysis. Commonly used models include Convolutional Neural Networks (CNNs) \cite{socher2011semi}, Recursive Neural Network (ReNNs) \cite{socher2013recursive}, and Recurrent Neural Networks (RNNs) \cite{irsoy2014opinion}. RNN naturally benefits sentiment classification because of its ability to capture sequential information in text. However, standard RNNs suffer from the so-called \textit{gradient vanishing problem} \cite{bengio1994learning} where gradients may grow or decay exponentially over long sequences. LSTM models are adopted to solve the gradient vanishing problem. An LSTM model provides a gated mechanism to keep the long-term memory. Each LSTM layer is generally followed by mean pooling and the output is fed into the next layer. Experiments in datasets which contain sentences and long documents demonstrate that LSTM model outperforms the traditional RNNs \cite{tang2015document,tang-qin-liu:2015:ACL-IJCNLP}. Attention mechanism is also added to LSTM models to highlight important segments at both sentence level and document level. Attention models can be built from text in local context \cite{yang2016hierarchical}, user/production information \cite{chen2016neural,long2017fake} and other information such as cognition grounded eye tracking data \cite{long2017cognition}. LSTM models with attention mechanism are currently the state-of-the-art models in document sentiment analysis tasks \cite{chen2016neural,long2017cognition}. 

Memory networks are designed to handle larger context for a collection of documents. Memory networks introduce inference components combined with a so called long-term memory component \cite{weston2014memory}. The long-term memory component is a large external memory to represent data as a collection. This collective information can contain local context \cite{das2017question} or external knowledge base \cite{jain2016question}. It can also be used to represent the context of users and products globally \cite{tang2016aspect}. Dou uses \shortcite{dou2017capturing}  a memory network model in document level sentiment analysis and makes comparable result to the state-of-the-art model \cite{chen2016neural}.

\subsection{Incorporating User and Product Information}\label{subsec:kbinsa} 
Both user profile and product information have crucial effects on sentiment polarities. Tang et al. \shortcite{tang2015learning} proposed a model by incorporating user and product information into a CNN network for document level sentiment classification. User ids and product names are included as features in a unified document vector using the vector space model such that document vectors capture important global clues include individual preferences and product information.

Nevertheless, this method suffers from high model complexity and only word-level preference is considered rather than information at the semantic level \cite{chen2016neural}. Gui et al. \shortcite{gui2016intersubjectivity} introduce an inter-subjectivity network to link users to the terms they used as well as the polarities of the terms. The network aims to learn writer embeddings which are subsequently incorporated into a CNN network for sentiment analysis. Chen et al. \shortcite{chen2016neural} propose a model to incorporate user and product information into an LSTM with attention mechanism. This model is reported to produce the state-of-the-art results in the three benchmark datasets (IMDB, Yelp13, and Yelp14). Dou \shortcite{dou2017capturing} also proposes a deep memory network to integrate user profile and product information in a unified model. However, the model only achieves a comparable result to the state-of-the-art attention based LSTM \cite{chen2016neural}.

\section{The DUPMN Model}\label{sec:proposaledmodel}
We propose a DUPMN model. Firstly, document representation is learned by a hierarchical LSTM network to obtain both sentence-level representation and document level representation  \cite{sundermeyer2012lstm}. A memory network model is then trained using dual memory networks, one for training user profiles and the other for training product reviews. Both of them are joined together to predict sentiment for documents.

\subsection{Task Definition}
Let $D$ be the set of review documents for classification, $U$ be the set of users, and $P$ be the set of products. For each document $d$($d \in D$), user $u$($u \in U$) is the writer of $d$ on product $p$($p \in P$). Let $U_{u}(d)$ be all documents posted by $u$ and $P_{p}(d)$ be all documents on $p$. $U_{u}(d)$ and $P_{p}(d)$ define the user context and the product context of $d$, respectively. For simplicity, we use $U(d)$ and $P(d)$ directly. The goal of a sentiment analysis task is to predict the sentiment label for each $d$. 

\subsection{Document Embedding}
    Since review documents for sentiment classification such as restaurant reviews and movie comments are normally very long, a proper method to embed the documents is needed to speed up the training process and achieve better accuracy. Inspired by the work of Chen \cite{chen2016neural}, a hierarchical LSTM network is used to obtain embedding representation of documents. The first LSTM layer is used to obtain sentence representation by the hidden state of an LSTM network. The same mechanism is also used for document level representation with sentence-level representation as input. User and product attentions are included in the network so that all salient features are included in document representation. For document $d$, its embedding is denoted as $\vec{d}$. $\vec{d}$ is a vector representation with dimension size $n$. In principle, the embedding representation of user context of $d$, denoted by $\hat{U}(d)$,  and product context $\hat{P}(d)$ vary depending on $d$. For easy matrix calculation, we take $m$ as our model parameter so that $\hat{U}(d)$ and $\hat{P}(d)$ are two fixed $n \times m$ matrices.

\subsection{Memory Network Structure}
\begin{figure*}[h]
\includegraphics[width=\textwidth]{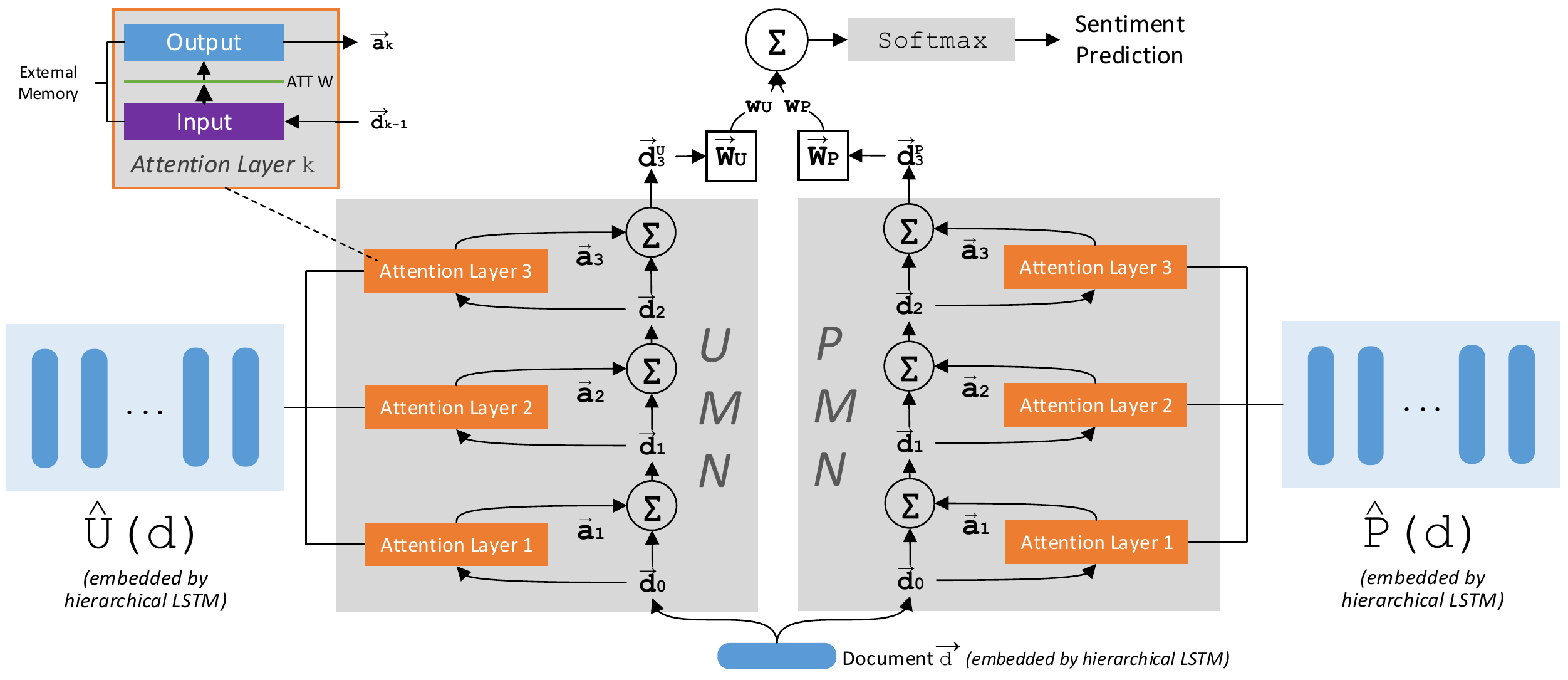}
\caption{Structure for Proposed DUPMN Model}
\label{fig:ours}
\end{figure*}

Inspired by the successful use of memory networks in language modeling, question answering,  and sentiment analysis \cite{sukhbaatar2015end, tang2016aspect,dou2017capturing}, we propose our DUPMN by extending a single memory network model to two memory networks to reflect different influences from users' perspective and products' perspective. The structure of the model is shown in Figure~\ref{fig:ours} with 3 hops as an example although in principle a memory network can have $K$ computational hops. 

The DUPMN model has two separate memory networks: the UMN and the PMN. Each hop in a memory network includes an attention layer $Attention_i$ and a linear addition $\Sigma_k$. Since the external memory $\hat{U}(d)$ and $\hat{P}(d)$ have the same structure, we use a generic notation $\hat{M}$ to denote them in the following explanations. Each document vector $\vec{d}$ is fed into the first hop of the two networks ($\vec{d}_0$=$\vec{d}$). Each $\vec{d}_{k-1}$( k= 1 ...... K-1) passes through the attention layer using an attention mechanism defined by a softmax function to obtain the attention weights $\vec{p}_k$ for document $d$:

\begin{equation} \label{eq:prob}
\vec{p}_k = Softmax(\vec{d}_{k-1}^T * \hat{M}),
\end{equation}
And to produce an attention weighted vector $\vec{a}_k$ by 
\begin{equation} \label{eq:output}
\vec{a}_k = \sum_{i=0}^{m}{{p_{k}}_{i}*\vec{M}_{i}}.
\end{equation}
$\vec{a}_k$ is then linearly added to $\vec{d}_{k-1}$ to produce the output of this hop as $\vec{d}_k$.

After completing the $K$th hop, the output $\vec{d}^u_K$ in UMN and $\vec{d}^p_K$ in PMN are joined together using a weighted mechanism to produce the output of DUPMN, $Output_{DUPMN}$, is given below: 
\begin{equation} 
{Output_{DUPMN}} = w_U{\vec{W}_U}{\vec{d}^u_K}+w_P{\vec{W}_P}{\vec{d}^p_K}.
\label{eq:combine2}
\end{equation}
Two different weight vectors $\vec{W}_u$ and $\vec{W}_p$ in Formula \ref{eq:combine2} can be trained for UMN and PMN. $w_U$ and $w_P$ are two constant weights to reflect the relative importance of user profile $\vec{d}^u_K$ and product information $\vec{d}^p_K$. 
The parameters in the model including $\vec{W}_U$, $\vec{W}_P$, $w_U$ and $w_P$. By minimizing the loss, those parameters can be optimized. 

Sentiment prediction is obtained through a $Softmax$ layer. The loss function is defined by the cross entropy between the prediction from $Output_{DUPMN}$ and the ground truth labels.

\section{Experiment and Result Analysis}\label{sec:Experiment}
Performance evaluations are conducted on three datasets and DUPMN is compared with a set of commonly used baseline methods including the state-of-the-art LSTM based method \cite{chen2016neural,wu2018improving}.

\subsection{Datasets}
\label{sec:Result and analysis}
The three benchmarking datasets include movie reviews from IMDB, restaurant reviews from Yelp13 and Yelp14 developed by Tang \shortcite{tang2015document}. All datasets are tokenized using the Stanford NLP tool \cite{manning2014stanford}. 
Table \ref{tablethreedatasets} lists statistics of the datasets including the number of classes, number of documents, average length of sentences, the average number of documents per user, and the average number of documents per product. Since postings in social networks by both users and products follow the long tail distribution \cite{kordumova2016exploring}, we only show the distribution of total number of posts for different products. For example, \#p(0-50) means the number of products which have reviews between the size of 0 to 50. We split train/development/test sets at the rate of 8:1:1 following the same setting in \cite{tang2015learning,chen2016neural}. The best configuration by the development dataset is used for the test set to obtain the final result. 
\begin{table}[]
\centering
\begin{tabular}{l"r|r|r}
\hline
        & IMDB      & Yelp13    & Yelp14    \\ \hline
\#class & 10        & 5         & 5         \\
\#doc   & 84,919    & 78,966    & 231,163   \\
\#users  & 1,310     & 1,631     & 4,818     \\
\#products   & 1,635     & 1,631     & 4,194     \\
Av sen. len   & 24.56     & 17.37     & 17.25     \\
Av docs/user   & 64.82     & 48.41     & 47.97     \\
Av docs/prod  & 51.93     & 48.41     & 55.12     \\  \hline
\#p(0-50)    & 1,223 & 1,299 & 3,150 \\ 
\#p(50-100)  & 318   &254   & 749   \\
\#p(100-150) & 72     & 56     & 175   \\
\#p(150-200) & 22     & 24     & 120  \\ \hline
\end{tabular}
\caption{Statistics of the three benchmark datasets}
\label{tablethreedatasets}
\end{table}

\begin{table*}[tbp]
\centering
\begin{tabular}{l|l"r"r|r|r|r|r|r|r|r}
\hline
                                 &        & \multicolumn{3}{c|}{IMDB} & \multicolumn{3}{c|}{Yelp13} & \multicolumn{3}{c}{Yelp14} \\ \hline
                        & Model            & Acc    & RMSE   & MAE    & Acc     & RMSE    & MAE    & Acc     & RMSE    & MAE    \\ \hline
\multirow{5}{*}{G1} & Majority         & 0.196  & 2.495  & 1.838  & 0.392   & 1.097   & 0.779  & 0.411   & 1.060    & 0.744  \\ \cline{2-11}
                        & Trigram          & 0.399  & 1.783  & 1.147  & 0.577   & 0.804   & 0.487  & 0.569   & 0.814   & 0.513  \\ \cline{2-11}
                        & TextFeature      & 0.402  & 1.793  & 1.134  & 0.572   & 0.800    & 0.490   & 0.556   & 0.845   & 0.520   \\ \cline{2-11}
                        & AvgWordvec & 0.304  & 1.985  & 1.361  & 0.530    & 0.893   & 0.562  & 0.526   & 0.898   & 0.568  \\ \hline \hline
\multirow{5}{*}{G2} & SSWE       & 0.312    & 1.973  & N/A  & 0.549   & 0.849   & N/A & 0.557   & 0.851   & N/A  \\ \cline{2-11}
                        & RNTN+RNN           & 0.400  & 1.734  & N/A  & 0.574   & 0.804  & N/A  & 0.582   & 0.821   & N/A  \\ \cline{2-11}
                         & CLSTM        & 0.421 & 1.549  & N/A & 0.592  & 0.729    & N/A & 0.637   & 0.686   & N/A  \\ \cline{2-11}
                        & LSTM+LA         & 0.443  & 1.465  & N/A  & 0.627   & 0.701   & N/A  & 0.637   & 0.686   & N/A  \\ \cline{2-11}
                        & LSTM+CBA       & 0.489  & 1.365  & N/A  & 0.638   & 0.697   & N/A  & 0.641   & 0.678   & N/A  \\ \hline \hline
\multirow{3}{*}{G3} 
& UPNN & 0.435  & 1.602  & 0.979  & 0.608   & 0.764    & 0.447  & 0.596   & 0.784   & 0.464  \\ \cline{2-11}
& UPDMN& 0.465  & 1.351  & 0.853  & 0.613   & 0.720    & 0.425  & 0.639   & 0.662   & 0.369  \\ \cline{2-11}
                     & InterSub     & 0.476 & 1.392   & N/A  & 0.623   & 0.714    & N/A  & 0.635   & 0.690   & N/A  \\ \cline{2-11}
                     & LSTM+UPA& \underline{0.533}  & \underline{1.281}  & N/A  & \underline{
0.650}   & \underline{0.692}   & N/A  & \underline{0.667}   & \underline{0.654}   & N/A  \\  \hline \hline
\multirow{1}{*}{New}               & DUPMN             & \textbf{0.539}  & \textbf{1.279}  & \textbf{0.734}  & \textbf{0.662}   & \textbf{0.667}   & \textbf{0.375}  & \textbf{0.676}    & \textbf{0.639}   & \textbf{0.351}\\ \hline \hline
\end{tabular}
\caption{Evaluation of different methods; best result/group in accuracy is marked in bold; second best is underlined.}
\label{tablecom3group}
\end{table*}

\subsection{Baseline Methods}\label{subsec:overallcp}
In order to make a systematic comparison, three groups of baselines are used in the evaluation. Group 1 includes all commonly used feature sets mentioned in Chen et al. \shortcite{chen2016neural} including Majority, Trigram, Text features (TextFeatures), and AveWordvec. All feature sets in Group 1 except Majority use the SVM classifier.

Group 2 methods include the recently published sentiment analysis models which only use context information, including: 
\begin{itemize}
\item \textbf{SSWE} \cite{tang2014learning} --- An SVM model using sentiment specific word embedding.
\item \textbf{RNTN+RNN} \cite{socher2013recursive} --- A Recursive Neural Tensor Network (RNTN) to represent sentences.
\item \textbf{CLSTM} \cite{xu2016cached} --- A Cached LSTM model to capture overall semantic information in long text. 
\item \textbf{LSTM+LA} \cite{chen2016neural} --- A state-of-the-art LSTM using local context as attention mechanism at both sentence level and  document level.
\item \textbf{LSTM+CBA} \cite{long2017cognition}--- A state-of-the-art LSTM model using cognition based data to build attention mechanism. 
\end{itemize}

Group 3 methods are recently published neural network models which incorporate user and product information, including:
\begin{itemize}
\item \textbf{UPNN} \cite{tang2015learning} --- User and product information for sentiment classification at document level based on a CNN network.
\item \textbf{UPDMN} \cite{dou2017capturing} --- A deep memory network for document level sentiment classification by including user and product information in a unified model. Hop 1 gives the best result, and  thus K=1 is used.
\item \textbf{InterSub} \cite{gui2016intersubjectivity} --- A CNN model making use of user and product information.
\item \textbf{LSTM+UPA} \cite{chen2016neural} --- The state-of-the-art LSTM including both local context based attentions and user/product in the attention mechanism.
\end{itemize}

For the DUPMN model, we also include two variations which use only one memory network. The first variation only includes user profiles in the memory network, denoted as \textbf{DUPMN-U}. The second variation only uses product information, denoted as \textbf{DUPMN-P}.

\subsection{Performance Evaluation}
Four sets of experiments are conducted. The first experiment compares DUPMN with other sentiment analysis methods. The second experiment evaluates the effectiveness of different hop size $K$ of memory network. The third experiment evaluates the effectiveness of UMN and PMN in different datasets. The fourth set of experiment examines the effect of memory size $m$ on the performance of DUPMN. Performance measures include Accuracy (ACC), Root-Mean-Square-Error (RMSE), and Mean Absolute Error (MAE) for our model. For other baseline methods in Group 2 and Group 3, their reported results are used. We also show the p-value by comparing the result of 10 random tests for both our model and the state-of-the-art model \footnote{We re-run experiment based on their public available code on GitHub (https://github.com/thunlp/NSC).} in the t-test \footnote{http://www.statisticshowto.com/probability-and-statistics/t-test/}.

\subsubsection*{Compared to other state-of-the-art models}
Table \ref{tablecom3group} shows the result of the first experiment. DUPMN uses one hop (the best performer) with $m$ being set at 100, a commonly used memory size for memory networks.

Generally speaking, Group 2 performs better than Group 1. This is because Group 1 uses a traditional SVM with feature engineering \cite{chang2011libsvm} and Group 2 uses more advanced deep learning methods proven to be effective by recent studies \cite{kim2014convolutional,chen2016neural}. However, some feature engineering methods are no worse than some deep learning methods. For example, the TextFeature model outperforms SSWE by a significant margin. 

When comparing Group 2 and Group 3 methods, we can see that user profiles and product information can improve performance as most of the methods in Group 3 perform better than methods in Group 2. This is more obvious in the IMDB dataset which naturally contains more subjectivity. In the IMDB dataset, almost all models with user and product information outperform the text-only models in Group 2 except LSTM+CBA \cite{long2017cognition}. However, the two LSTM models in Group 2 which include local attention mechanism do show that attention base methods can outperform methods using user profile and product information. In fact, the LSTM+CBA model using attention mechanism based on cognition grounded eye-tracking data in Group 2 outperforms quite a number of methods in Group 3. LSTM+CBA in Group 2 is only inferior to LSTM+UPA in Group 3 because of the additional user profile and production information used in LSTM+UPA. 

Most importantly, the DUPMN model with both user memory and product memory significantly outperforms all the baseline methods including the state-of-the-art LSTM+UPA model \cite{chen2016neural}. By using user profiles and product information in memory networks, DUPMN outperforms LSTM+UPA in all three datasets. In the IMDB dataset, our model makes 0.6 \% improvement over LSTM+UPA in accuracy with $p-value$ of 0.007. Our model also achieves lower RMSE value. In the Yelp review dataset, the improvement is even more significant. DUPMN  achieves 1.2\% improvement in accuracy in Yelp13 with $p-value$ of 0.004 and 0.9\% in Yelp14 with $p-value$ of 0.001, and the lower RMSE obtained by DUPMN also indicates that the proposed model can predict review ratings more accurately.

\begin{table*}[ht]            
\centering
                \resizebox{\textwidth}{!}{
                \begin{tabular}{l|c|c|c|c|c|c|c|c|c}
                \hline
                \multirow{2}{*}{}   & \multicolumn{3}{c|}{IMDB} & \multicolumn{3}{c|}{Yelp13} & \multicolumn{3}{c}{Yelp14} \\ \cline{2-10} 
                & \small{Acc}         & \small{RMSE}   & \small{MAE}     & \small{Acc}& \small{RMSE} & \small{MAE}& \small{Acc} & \small{RMSE}& \small{MAE}    \\ \hline \hline
                \footnotesize{DUPMN-U(1)}        & \underline{0.536}        & \underline{1.273}   & \underline{0.737}     &0.656    & 0.687 & 0.380       & 0.667        & 0.655  & 0.361   \\ \hline               
                \footnotesize{DUPMN-U(2)}        & 0.526       & 1.285   & 0.748     & 0.653     & 0.689 & 0.382       & 0.665          & 0.661 & 0.369      \\ \hline
                \footnotesize{DUPMN-U(3)}  & 0.524         & 1.295   & 0.754     & 0.651      & 0.692 & 0.388       & 0.661         & 0.667  & 0.374      \\ \hline \hline
                \footnotesize{DUPMN-P(1)}  & 0.523        & 1.346   & 0.769     & \underline{0.660}     & \underline{0.668} & \underline{0.370}       & \underline{0.670}         & \underline{0.649}  & \underline{0.357}          \\ \hline
                \footnotesize{DUPMN-P(2)}       & 0.517         & 1.348   & 0.775     &0.656     & 0.680 & 0.380       & 0.667          & 0.656  & 0.364  \\ \hline
                \footnotesize{DUPMN-P(3)}                        & 0.512         & 1.356   & 0.661     & 0.651    & 0.699 & 0.388       & 0.661         & 0.661  & 0.370   \\  \hline \hline
                \footnotesize{DUPMN(1)}     & \textbf{0.539}         & \textbf{1.279}   & \textbf{0.734}     & \textbf{0.662}     & \textbf{0.667}  &\textbf{0.375} & \textbf{0.676}          & \textbf{0.639}  & \textbf{0.351}  \\ \hline
                \footnotesize{DUPMN(2)}     & 0.522        & 1.299   & 0.758     & 0.650     & 0.700 & 0.390       & 0.667          & 0.650  & 0.359  \\ \hline
                \footnotesize{DUPMN(3)}     & 0.502         & 1.431   & 0.830     & 0.653      & 0.686 & 0.382       & 0.658          & 0.668  & 0.371  \\ \hline
                \hline
                \end{tabular}}
                \caption{Evaluation of different memory network hops and user and product information utilization\protect\footnotemark}
                \label{taballbranches}
\end{table*}
            \footnotetext{Best results are marked in bold; second best are underlined in the table}

\subsubsection*{Effects of different hop sizes}
The second set of experiments evaluates the effectiveness of DUPMN using different number of hops $K$. Table \ref{taballbranches} shows the evaluation results. The number in the brackets after each model name indicates the number of hops used. Two conclusions can be obtained from Table \ref{taballbranches}. We find that more hops do not bring benefit. In all the three models, the single hop model obtains the best performance. Unlike video and image information, written text is grammatically structured and contains abstract information such that multiple hops may introduce more information distortion. Another reason may be due to over-fitting by the additional hops. 

\begin{table}[ht]
\centering
\begin{tabular}{c|c|c|c|c|c}
\hline
\multicolumn{2}{c|}{IMDB} & \multicolumn{2}{c|}{Yelp13} & \multicolumn{2}{c}{Yelp14} \\ \cline{1-6}
        			\small{$w_U$}&\small{$w_P$}&\small{$w_U$}&\small{$w_P$}&\small{$w_U$}&\small{$w_P$} \\ \hline \hline
                    \textbf{0.534} & 0.466 & 0.475 & \textbf{0.525} & 0.436 & \textbf{0.564} \\ \hline 
                    \end{tabular}
                    \caption{Average weight of UMN and PMN in different datasets}
                    \label{tab:combine_weight}
\end{table}

\subsubsection*{Effects of DUPMN-U and DUPMN-P}
Comparing the performance of DUPMN-U and DUPMN-P in Table \ref{taballbranches}, it also shows that user memory and product memory indeed provide different kinds of information and thus their usefulness are different in different datasets. For the movie review dataset, IMDB, which is more subjective, results show that user profile information using DUPMN-U outperforms DUPMN-P as there is a 1.3\% gain compared to that of DUPMN-P. However, on restaurant reviews in Yelp datasets, DUPMN-P performs better than DUPMN-U indicating product information is more valuable.

To further examine the effects of UMN and PMN on sentiment classification, we observe the difference of optimized values of the constant weights $w_{U}$ and $w_{P}$ between the UMN and the PMN given in Formula \ref{eq:combine2}. The difference in their values indicates the relative importance of the two networks. The optimized weights given in Table \ref{tab:combine_weight} on the three datasets show that user profile has a higher weight than product information in IMDB because movie review is more related to personal preferences whereas product information has a higher weight in the two restaurant review datasets. This result is consistent with the evaluation in Table \ref{taballbranches} on DUPMN-U and DUPMN-P.

\begin{figure}[!h]
\includegraphics[width=\linewidth]{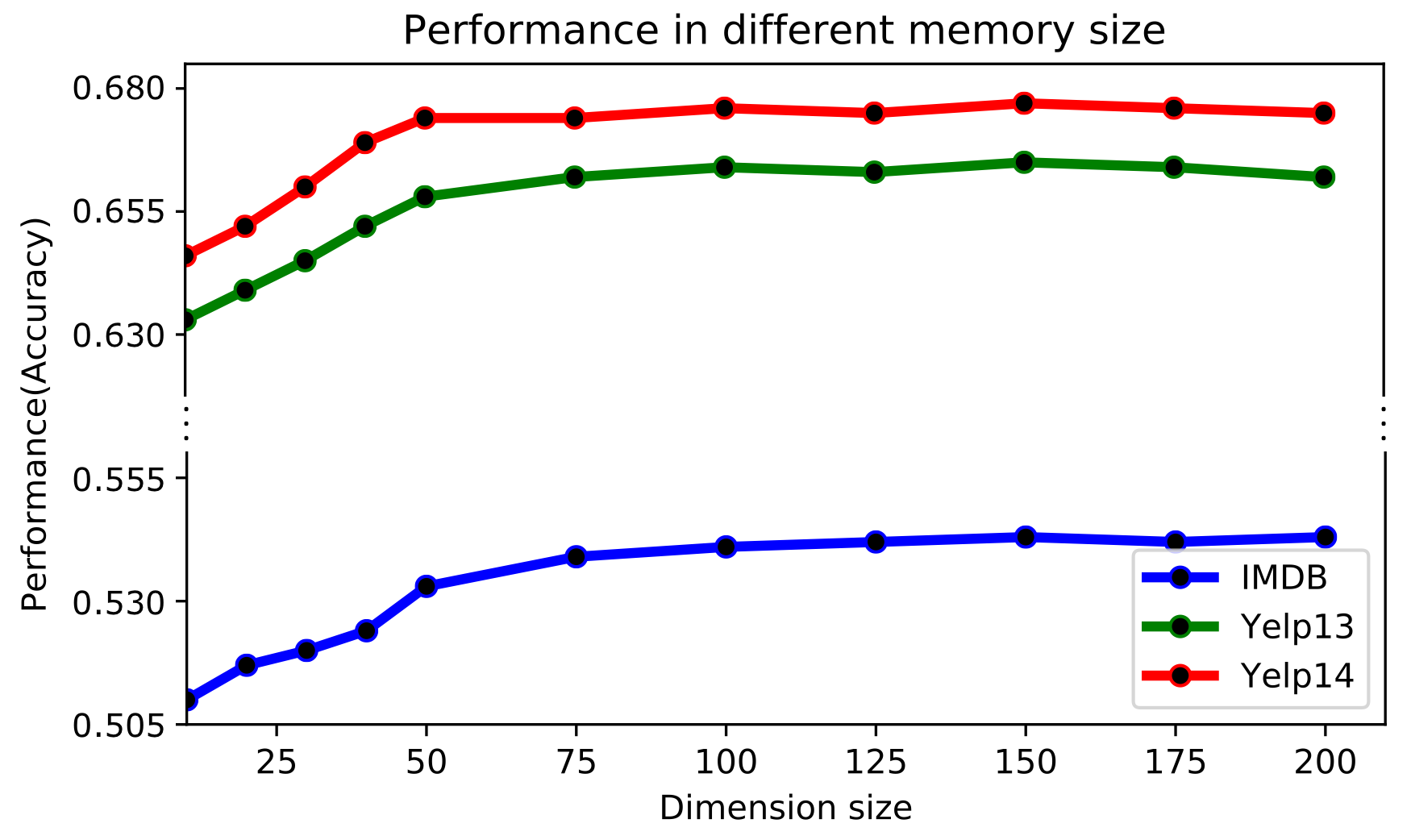}
\caption{Effect of different memory sizes}
\label{fig:dimensize}
\end{figure}

\subsubsection*{Effects of the memory size}
Most social network data follows the long tail distribution. If the memory size to represent the data is too small, some context information will be lost. On the other hand, too large memory size which requires more resources in computation and storage may not introduce much benefit. Thus, the fourth set of experiments evaluates the effect of dimension size $m$ in the DUPMN memory networks. Figure \ref{fig:dimensize} shows the result of the evaluation for 1 hop configuration with memory size starting at 1 with 10 points at each increment until size of 75, the increment set to 25 from 75 to 200 to cover most postings. Results show that when memory size increases from 10 to 100, the performance of DUPMN steadily increases. Once it goes beyond 100, DUPMN is no longer sensitive to memory size. This is related to the distribution of document frequency rated by user/product in Table \ref{tablethreedatasets} as the average is around 50. With long tail distribution, after 75, not many new documents will be included in the context. To improve algorithm efficiency without much compromise on performance, $m$ can be any value that doubles the average. So, values between 100-200 in our algorithm should be quite sufficient.

\subsection{Case Analysis}
The review text below is for a sci-fi movie which has the golden label 10 (most positive). However, if it is read as an isolated piece of text, identifying its sentiment is difficult. The LSTM+LA model gives it the rating of 1 (most negative), perhaps because on the surface, there are many negative words like \textit{unacceptable}, \textit{criticize} and \textit{sucks} even though the reviewer is praising the movie.  Since our user memory can learn that the reviewer is a fan of sci-fi movies, our DUPMN model indeed gives the correct rating of 10.


\begin{small}
\textit{okay, there are two types of movie lovers: ... they expect to see a Titanic every time they go to the cinema ... this movie sucks? ... it is definitely better than other sci-fi ..... the audio and visual effects are simply terrific and Travolta's performance is brilliant-funny and interesting. what people expect from sci-fi is beyond me ... the rating for Battlefield Earth is below 2.5, which is unacceptable for a movie with such craftsmanship. Scary movie, possibly the worst of all time - ..., has a 6! maybe we should all be a little more subtle when we criticize movies... especially sci-fi.., since they have become an endangered genre ... give this movie the recognition it deserves.}
\end{small}

\section{Conclusion and Future Work}\label{sec:Conclusion and future works}
We propose a novel dual memory network model for sentiment predictions.
We argue that user profile and product information are fundamentally different as user profiles reflect more on subjectivity whereas product information reflects more on salient features of products at aggregated level. Based on this hypothesis, two separate memory networks for user context and product context are built 
at the document level through a hierarchical learning model. The inclusion of an attention layer can further capture semantic information more effectively. Evaluation on three benchmark review datasets shows that the proposed DUPMN model outperforms the current state-of-the-art systems with significant improvements shown in p-value of 0.007, 0.004 and 0.001 respectively. We also show that single hop memory networks is the most effective model.  Evaluation results show that user profile and product information are indeed different and have different effects on different datasets. 
In more subjective datasets such as IMDB, the inclusion of user profile information is more important. Whereas on more objective datasets such as Yelp data, collective information of restaurant plays a more important role in classification.

Future works include two directions. One direction is to explore the contribution of user profiles and product information in aspects level sentiment analysis tasks. Another direction is to explore how knowledge-based information can be incorporated to further improve sentiment classification tasks. 

\section*{Acknowledgments}
The work is partially supported by the research grants from Hong Kong Polytechnic University (PolyU RTVU) and GRF grant (CERG PolyU 15211/14E, PolyU 152006/16E).



\bibliography{emnlp2018}
\bibliographystyle{acl_natbib_nourl}

\end{document}